\documentclass{article}
\pdfoutput=1
\usepackage[bookmarks=false]{hyperref}

\usepackage[preprint]{neurips_2024}

\usepackage[utf8]{inputenc} 
\usepackage[T1]{fontenc}    
\usepackage{hyperref}       
\usepackage{url}            
\usepackage{booktabs}       
\usepackage{amsfonts}       
\usepackage{nicefrac}       
\usepackage{microtype}      
\usepackage{xcolor}         

\usepackage{multirow}
\usepackage{multirow}
\usepackage{graphicx}
\usepackage{wrapfig}
\usepackage{svg}
\usepackage{amsmath}
\usepackage{makecell}

\DeclareMathOperator{\MLP}{MLP}

\DeclareMathOperator{\gumbelsoftmax}{Gumbel-Softmax}

\DeclareMathOperator{\GELU}{GELU}

\DeclareMathOperator{\mse}{MSE}
\DeclareMathOperator{\crossentropy}{CrossEntropy}

\title{Saliency-driven Dynamic Token Pruning for Large Language Models}

\author{
  Yao Tao, Yehui Tang, Yun Wang, Mingjian Zhu, \\
  \bf{Hailin Hu\thanks{Corresponding Author.}, Yunhe Wang$^{\ast}$} \\
  \small Huawei Noah's Ark Lab. \\
  \small\texttt{\{hailin.hu,yunhe.wang\}@huawei.com}\\
}

\begin{document}

\maketitle

\begin{abstract}
  Despite the recent success of large language models (LLMs), LLMs are particularly challenging in long-sequence inference scenarios due to the quadratic computational complexity of the attention mechanism. Inspired by the interpretability theory of feature attribution in neural network models, we observe that not all tokens have the same contribution. Based on this observation, we propose a novel token pruning framework, namely Saliency-driven Dynamic Token Pruning (SDTP), to gradually and dynamically prune redundant tokens based on the input context. Specifically, a lightweight saliency-driven prediction module is designed to estimate the importance score of each token with its hidden state, which is added to different layers of the LLM to hierarchically prune redundant tokens. Furthermore, a ranking-based optimization strategy is proposed to minimize the ranking divergence of the saliency score and the predicted importance score. Extensive experiments have shown that our framework is generalizable to various models and datasets. By hierarchically pruning 65\% of the input tokens, our method greatly reduces 33\% $\sim$ 47\% FLOPs and achieves speedup up to 1.75$\times$ during inference, while maintaining comparable performance. We further demonstrate that SDTP can be combined with KV cache compression method for further compression.

\end{abstract}

\section{Introduction}
Due to the quadratic complexity of the attention mechanism, the efficiency of a large language model is highly influenced by extended input and output sequence length. Using Mistral-7B as an example, a model with 7 billion parameters. When the input sequence length is increased from 4K tokens to 128K tokens, the TFLOPs required for processing surges from 72.51 to 8864.49, reflecting a 122-fold rise. This significant growth highlights the substantial computational demands imposed by longer sequences on large language models. To overcome this limitation, token or context pruning turns out to be an important research aspect. On the one hand, precise pruning of less informational tokens directly reduces the input size and the computation FLOPs involved. On the other hand, different from neural network pruning~\cite{liu2023deja, fan2019reducing, ma2023llm, frantar2023sparsegpt}, distillation~\cite{gu2023minillm, jiao2019tinybert, sanh2019distilbert} or quantization~\cite{frantar2022gptq, xiao2023smoothquant, wu2023understanding}, it avoids the modification of the original model deployment, therefore the latency improvement is guaranteed across different hardware deployments~\cite{rao2021dynamicvit}. 

Despite the conceptual advancement, the current token pruning methods still face challenges in LLM inference. Firstly, most methods focus on KV cache compression during autoregressive decoding~\cite{zhang2024h2o, liu2024scissorhands, anagnostidis2024dynamic, xiao2023efficient}. While these methods are generally beneficial to LLM inference speed, they are not designed for the computation-bounded stage of prefill, leading to a sub-optimal acceleration in many real world scenarios, especially when prompt engineering generates a long in-context content injection for the input~\cite{jiang2023llmlingua, pan2024llmlingua}. Moreover, the token selection criteria is often limited to local attention statics (e.g.,~\cite{zhang2024h2o}) or sigmoid functions (e.g.,~\cite{anagnostidis2024dynamic}), which may neglect important semantic information in the pruning process. 

\begin{figure}[t]
  \centering
  \begin{minipage}[b]{0.63\textwidth}
  \centering
    \includegraphics[height=2.7cm]{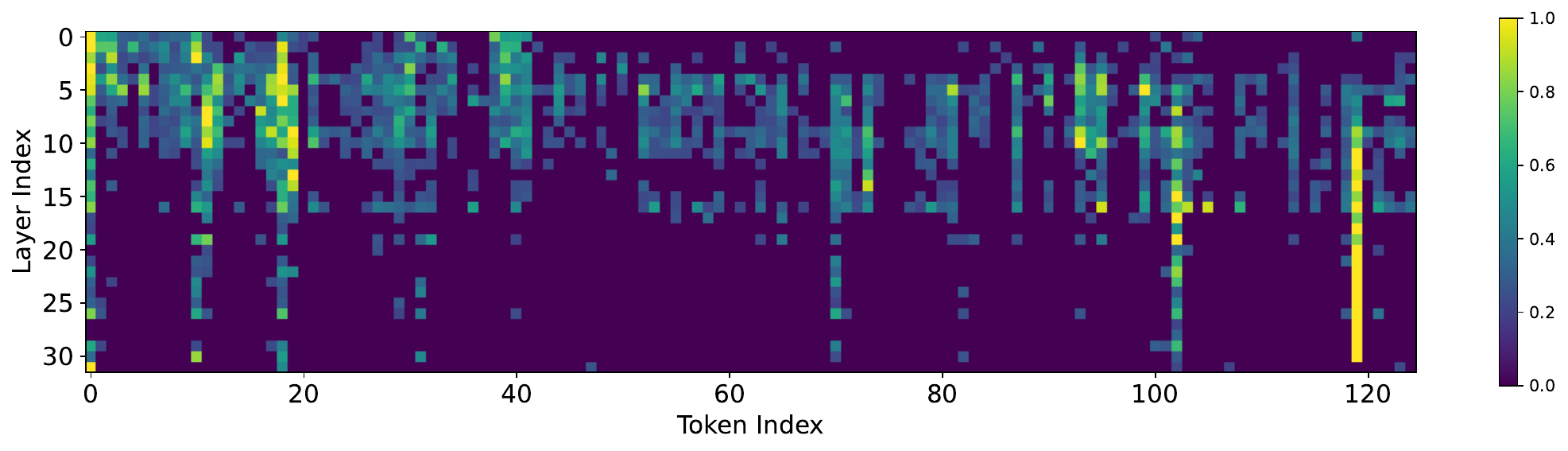}
    \caption{Gradient-based saliency scores.}
    \label{fig:salient_token_sparsity_1}
  \end{minipage}
  \hfill
  \begin{minipage}[b]{0.35\textwidth}
  \centering
    \includegraphics[height=2.7cm]{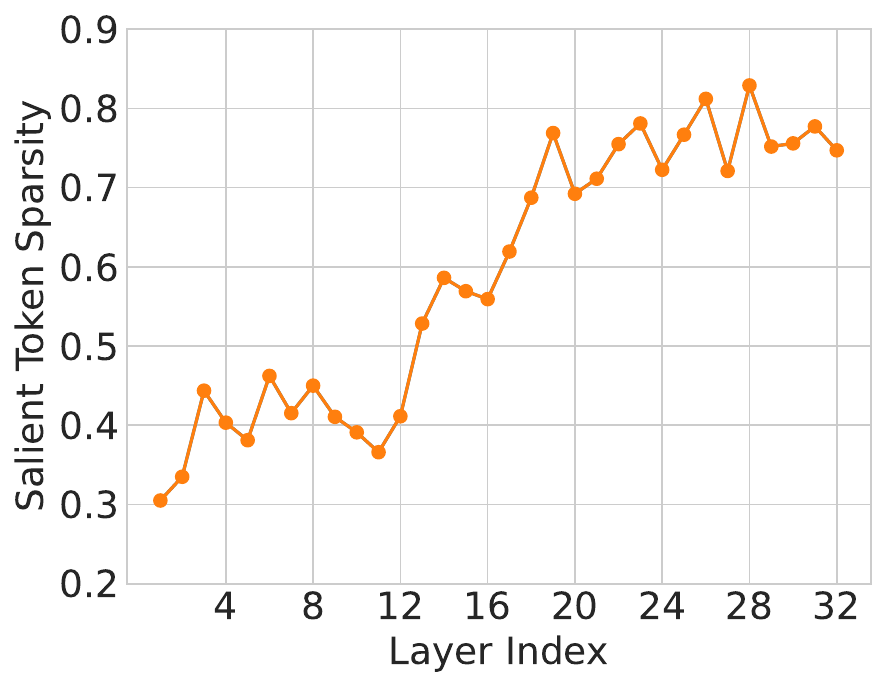}
    \caption{Salient token sparsity. }
    \label{fig:salient_token_sparsity}
  \end{minipage}
\end{figure}

To overcome these limitations, in this study, we propose a novel token pruning framework, namely Saliency-driven Dynamic Token Pruning (SDTP), to perform a cascade pruning process across the Transformer layers to speed up inference. The motivation of our method originates from gradient-based model interpretability~\cite{kindermans2019reliability}. Following the intuition that tokens with higher gradient-based saliency scores indicate more linguistic contribution, we first conduct a preliminary analysis of observations on the saliency map of LLM models. Figure~\ref{fig:salient_token_sparsity_1} shows a typical example using LIama2-7B, where we find that not all input tokens have the same contribution. In particular, we find that a) important tokens are sparse, with a sparse rate increasing with the number of layers; and b) the sparsity can largely be passed to deeper layers, i.e., if the token at the front layer is determined as redundant, it will still be redundant in deeper layers. The sparsity pattern can also be statistically verified in Figure~\ref{fig:salient_token_sparsity} using the saliency statistics on the Dolly dataset. The important tokens are defined as the tokens whose saliency score is greater than 10\% of the maximum value, and the rest are redundant tokens.

Our framework focuses on determining the most informative tokens given the input prompt. In particular, we leverage a gradient-based method to dynamically mark important tokens across the layers, and finally generate a pruned input before the first token generation. Our main contributions are summarized as follows:
\begin{itemize}
\item We develop a token selection scheme by pruning the input length, which benefits the whole process of LLM inference, achieving nearly lossless speedup even with a high pruning ratio.
\item We propose a novel learnable pruning module supervised by gradient-based feature attribution and a ranking-based optimization strategy.
\item Extensive experiments have shown our framework is generalizable to various models and datasets by hierarchically pruning 65\% of the input tokens, achieving up to 47.2\% FLOPs reduction, 1.75$\times$ speeding up, and 34.26\% memory savings for Mistral-7B.

\end{itemize}

\section{Related Work}
\paragraph{Structure pruning for transformer.} In the context of transformer models for natural language processing tasks, a significant body of work has concentrated on reducing the number of heads in the multi-head attention (MSA) module. For instance, Michel et al.~\cite{michel2019sixteen} observe that eliminating a large percentage of heads in pre-trained BERT models~\cite{devlin2018bert} has a minimal impact on performance. Beyond the MSA module, pruning has also been applied to the tokens, as demonstrated in~\cite{goyal2020power, kim2020length, wang2021spatten}. LTP~\cite{kim2022learned} is a threshold-based token pruning method that needs a simple threshold operation and fully automates the search for optimal pruning configurations with a differentiable soft binarized mask. Yun~\cite{yun2023focus} proposes an approach of integrating token pruning and token combining to improve document classification for the BERT model, which includes fuzzy-based token pruned attention and a token combining module that gradually removes unimportant tokens and reduces time and memory costs. 
\paragraph{Structure pruning for LLMs.} Recent approaches~\cite{fan2019reducing, kurtic2023ziplm, liu2024scissorhands} employ structural pruning methods on large language models (LLMs), removing entire structured components to achieve more efficient GPU speedups. Based on the contextual sparsity hypothesis, Deja Vu~\cite{liu2023deja} trims specific attention heads and MLP parameters without modifying pre-trained models. H2O~\cite{zhang2024h2o} is a KV cache eviction policy that identifies a subset of tokens which contribute the most value to attention scores. This method reduces memory footprint and improves LLM deployment efficiency, enhancing throughput and reducing latency without compromising generation quality. LLMLingua~\cite{jiang2023llmlingua} proposes a data distillation method to compress prompts from GPT-4~\cite{achiam2023gpt} by treating prompt compression as a token classification task, capturing essential information, reducing latency, and ensuring the compressed prompt's faithfulness to the original content. Unlike them, our proposed method investigates redundancy from a novel perspective by considering the information integration of different tokens in a transformer. Additionally, reducing tokens can be combined with pruning in other dimensions to achieve a faster acceleration.
\section{Method}
\begin{figure}[t]
	\centering
	\includegraphics[width=\textwidth]{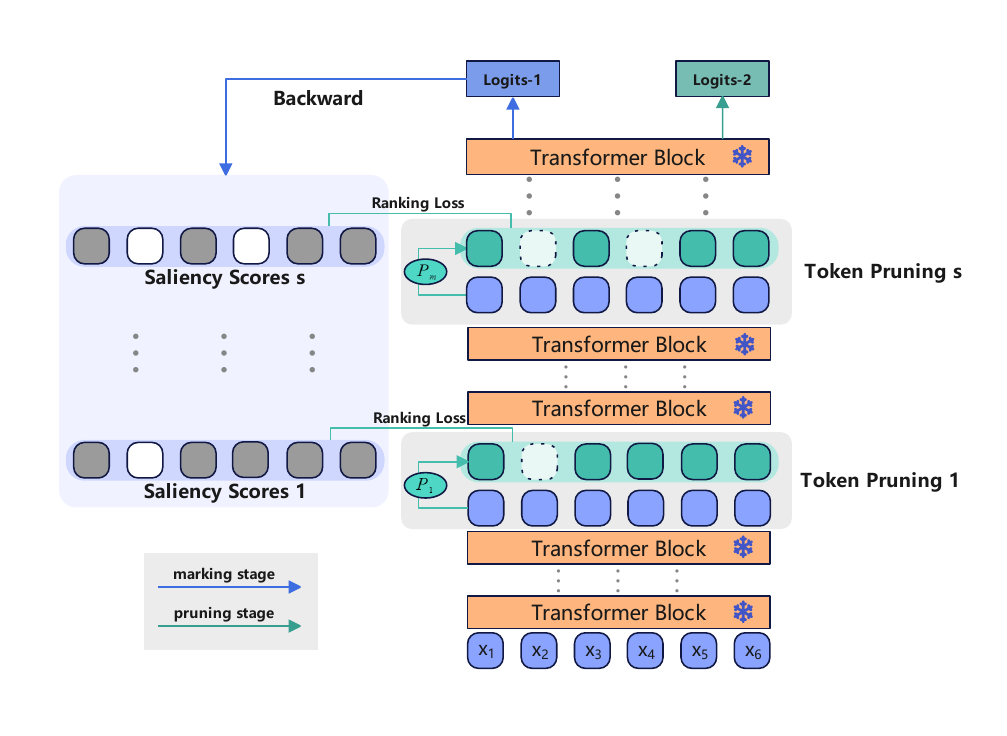}
	\caption{The overall architecture of our proposed method. Our proposed token pruning module is inserted between the transformer blocks and learns to decide the tokens to be pruned. The pruned tokens lead to less computation demand in the following layers.} 
	\label{fig:idea}
\end{figure}
\subsection{Overview}
Our method progressively identifies and removes non-informative tokens from the model's memory during inference. We achieve this by introducing learnable pruning modules within the Transformer architecture. These modules dynamically analyze token representations and select tokens for pruning. Each pruning decision leverages a corresponding gradient-based signal, guided by a dual-loss strategy detailed later. This strategy balances the token pruning objective with the preservation of the model's pre-trained capabilities through a language modeling loss. The pruning mask, indicating the tokens to be removed, is propagated through subsequent layers. Finally, all identified non-informative tokens are eliminated from memory before calculating the first token. This selective removal significantly reduces computational demands during inference, as the pruning module operates with a fraction of the FLOPs required by the entire language model (less than 1\%).
\subsection{Learnable Token Pruning Module}

The architecture of our SDTP is shown in Figure~\ref{fig:idea}. Our SDTP contains a typical transformer and a few token pruning modules. The transformer can be implemented as various large language models such as Llama2~\cite{touvron2023llama}, BLOOM~\cite{le2023bloom} and Mistral~\cite{jiang2023mistral}. The token pruning module decides which tokens are dropped by producing the probabilities for all the tokens. Tokens with a smaller score are dropped. We hierarchically insert the token pruning module through the transformer network at certain locations. Thus, the token sparsification is only conducted in these locations. We adopt a two-stage approach in training. The first stage is the token marking stage (blue line), using the original model and all the tokens to perform inference and compute the saliency scores using the output of the transformer. The saliency scores are further utilized in the second stage for computing losses. The second stage is called the token pruning stage (green line). The token pruning module takes effect. We supervise the learning of the token pruning module using MSE loss and a proposed ranking loss. A cross-entropy loss is also utilized to train the transformer. The token pruning module and the transformer can be optimized end-to-end. During inference, the most important tokens are selected and kept according to a predefined ratio in each location, and the scores are computed by the token pruning module.

In the second stage, we use a binary decision mask $M$ to identify the tokens to be kept and pruned. $N$ is the number of tokens. All elements in the decision mask are firstly set to 1 and we update the mask progressively. The token pruning modules take the existing tokens $x \in \mathbb{R}^{N}$ as input and outputs
the decision mask. The token pruning module is implemented as a two-layer MLP:
\begin{align}
	\begin{aligned}
	\pi = \MLP(\GELU(\MLP(\mathbf{x}))),
	\end{aligned}
	\label{eq:loss_real}
\end{align}
\begin{align}
	\begin{aligned}
	M = \gumbelsoftmax(\pi),
	\end{aligned}
	\label{eq:loss_real}
\end{align}
where $\pi \in \mathbb{R}^{N\times 2}$ is the output of the token pruning module. $M\in \{0,1\}^N$ is a one-hot tensor output from Gumbel-Softmax. $M$ denotes the mask of tokens that need to be kept. Gumbel-Softmax is a differentiable function, which makes it possible to conduct end-to-end training. GELU is the activation function. 

\subsection{Token Pruning Supervision}
Gradient-based attribution techniques~\cite{kindermans2019reliability, mohebbi2021exploring} can be used to determine the importance of each input feature through the analysis of partial derivatives of the output with respect to each input dimension. The magnitude of the derivatives serves to reflect the sensitivity of the output with respect to alterations in the input. We use this method to identify the important tokens and further supervise the training of the token pruning module:

\begin{align}
	\begin{aligned}
		\hat \pi =  \frac{\partial T(x)}{\partial x} \cdot x,
	\end{aligned}
	\label{eq:saliency}
\end{align}

where $T(x)$ is the output of the entire network and $x$ denotes the input vector of the selected layer that has a token pruning module. $\hat \pi$ indicates the importance of tokens in the transformer. Since the saliency score $\hat \pi$ cannot be computed in the inference stage, we utilize a token pruning module to simulate it. We use MSE loss to minimize the difference between the predictions of the token pruning module and saliency score $\hat \pi$.
\begin{equation}
	\mathcal{L}_{\rm mse} = \mse\left(\mathbf{\pi} , \mathbf{\hat \pi}\right),
\end{equation}

The MSE loss ensures the value of $\mathbf{\pi}$ and $\mathbf{\hat \pi}$ become closer. Beyond the value proximity, a more significant consideration is the proximity in the ranking of token importance, since we keep the tokens with higher importance values and a predefined ratio of tokens with smaller importance values are pruned. To minimize the ranking divergence of $\mathbf{\pi}$ and $\hat \pi$, we propose a ranking-based token selection loss to supervise the learning of token pruning, as shown in Figure~\ref{fig:idea}. The total ranking loss $\mathcal{L}_{\rm r}$ is the sum of ranking loss in each stage. For simplicity, we formulate this process as follows: 
\begin{equation}
	\mathcal{L}_r= \sum_{s=1}^{S} \mathcal{L}_{r}^{(s)},
\end{equation}
where $S$ is the largest number of selected stages that need to be pruned in the transformer. The ranking loss in each layer can be formulated as follows:

\begin{equation}
	\mathcal{L}^{(s)}_{\rm r} (\pi, \hat \pi) = \sum_{i=1}^{N-1}  \sum_{j=i+1}^{N} log{(1+e^{-((\pi_{i}-\pi_{j})*sign(\hat\pi_{i} - \hat\pi_{j}))})},
\end{equation}
where $N$ is the total number of tokens. The cross-entropy loss is also utilized to train the entire transformer:
\begin{align}
	\begin{aligned}
		\mathcal{L}_{\rm cls} = \crossentropy(y, \hat y),
	\end{aligned}
	\label{eq:loss_cls}
\end{align}
where $\hat y$ is the output of the transformer and $y$ is the label. The total loss is formulated as follows:
\begin{align}
	\begin{aligned}
		\mathcal{L} = \mathcal{L}_{\rm cls} + \mathcal{L}_{\rm mse} + \mathcal{L}_{\rm r}.
	\end{aligned}
	\label{eq:loss_cls}
\end{align}
Our method will be trained end-to-end. We can also distill the knowledge from the original model to the pruned model in training, which improves the performance of the pruned model.

\section{Experiments}
In this section, we aim to demonstrate that SDTP, a lightweight saliency-driven dynamic token pruning framework, can reduce memory usage and latency in wall-clock while maintaining generation quality across a broad spectrum of domains and tasks. Experiments on different LLMs show that the proposed method can be applied to all current mainstream Transformer-based LLMs. Ablation experiments confirm the effectiveness of our proposed SDTP framework and the ingenious ranking-based optimization strategy. We further illustrate the compatibility of our approach with KV cache compression techniques.
\subsection{Experimental Setup}
\paragraph{Implementation details.}
We conducted experiments using three representative LLM models, including the Llama2-7B\footnote{https://llama.meta.com/}, Mistral-7B-Instruct-v0.2\footnote{https://mistral.ai/} and BLOOM-7B\footnote{https://bigscience.huggingface.co/blog/bloom}. Unless otherwise specified, we fix the number of pruning stages $S=10$ and apply the target keeping ratio $r$ as a geometric sequence $[r, r^2, ..., r^{10}]$ where $r=0.9$. 

During the training phase, the pre-trained model is used for initialization, the weight of the original model is frozen, and only the weight of the SDTP module is iteratively updated. This brings two benefits: on the one hand, the strong generalization capability of LLM on downstream tasks can be retained. On the other hand, we only need low-resource alignment training to achieve excellent performance. In addition, for the stability of training, we use attention masking to simulate token pruning, just like in DynamicVit~\cite{rao2021dynamicvit}. All models are trained on the databricks-dolly-15K\footnote{https://github.com/databrickslabs/dolly/tree/master} dataset for two epochs. 

During the inference phase, referring to StreamLLM~\cite{xiao2023efficient} and $\rm H_2O$~\cite{zhang2024h2o}, we always keep the first four initial tokens and the local 10\% of the tokens. We set the batch size to 1 to test the accuracy and speed. We use the same hardware for training and evaluation as this work~\cite{shao2024flexibly}.

\paragraph{Evaluation of generality.}
In order to evaluate the generality of our SDTP on different downstream tasks, we employed the lm-eval-harness~\cite{gao2021framework} to compare the accuracy before and after dynamic token pruning on several few-shot downstream tasks, including COPA~\cite{roemmele2011choice}, PIQA~\cite{bisk2020piqa}, Winogrande~\cite{sakaguchi2021winogrande}, MathQA~\cite{amini2019mathqa}, BoolQ~\cite{clark2019boolq}, CB~\cite{roberts2020much}, WiC~\cite{pilehvar2018wic}, and WSC~\cite{kocijan2020review}. The evaluation metrics for each task refer to lm-eval-harness~\cite{gao2021framework}.
\paragraph{Evaluation of long context.}
For long-context scenarios, we conduct experiments on LongBench~\cite{bai2023longbench} with Mistral-7B-Instruct-v0.2\footnote{https://mistral.ai/}, which achieves compelling performance in long-context scenarios. We evaluate our method and LLMLingua-2~\cite{pan2024llmlingua} on the English datasets of LongBench, including single-document QA, multi-document QA, summarization, few-shot learning and code generation tasks. We prune the model with 5 and 10 pruning stages, denoted as SDTP-low and SDTP in Table~\ref{tab:longbench}, respectively. The compression ratio is the number of original tokens divided by the number of compressed tokens. In our method, the average number of input tokens of all layers is taken as the number of compressed tokens. The evaluation metrics for each dataset refer to LongBench github\footnote{https://github.com/THUDM/LongBench}. We evaluate the metrics and inference latency with Huggingface's transformers 4.36.0, PyTorch 2.0.1, and FlashAttention2 2.3.3.
\subsection{Main Results}
\paragraph{Results on long context benchmarks.}
Considering that there is no related work to dynamically prune tokens in the middle layers during inference in the LLM scenario, we compared our method with the prompt pruning method LLMLingua-2~\cite{pan2024llmlingua} in the out-of-domain scenario, which is a strong baseline. As illustrated in Table~\ref{tab:longbench}, with the same token compression ratio of 1.61x, our method SDTP achieves higher performance and speedup ratio on LongBench than LLMLingua-2. The average performance of our method is 4.66 higher than LLMLingua-2, as SDTP performs significantly better on both Code and FewShot tasks. The speedup ratio gain is attributed to our lightweight pruning module, while LLMLingua-2 uses xlm-roberta-large and multilingual-BERT as prompt compressors. These results demonstrate that our method can prune less important tokens more efficiently. To reduce the performance gap between our method and the original input tokens, SDTP-low is pruned with a lower compression ratio of 1.32x, and the performance drop on every task is reduced to 1$\sim$2.37.
\begin{table}[htbp]
\caption{Evaluation with Mistral-7B on LongBench. SDTP-low is pruned with a lower compression ratio than SDTP. $1/\tau$ is the compression ratio. * denotes our implementation based on the official codes.}
\label{tab:longbench}
\centering
\resizebox{\linewidth}{!}{
\begin{tabular}{l|cccccc|c|cc}
\toprule
\multirow{2}{*}{Methods} & \multicolumn{6}{c|}{LongBench} & \multirow{2}{*}{$1/\tau$}                         & \multicolumn{2}{c}{Speedup} \\ 
\cmidrule{2-7} \cmidrule{9-10} 
            & SingleDoc & MultiDoc & Summ. & FewShot & Code  & AVG &     & Prefill & End2End     \\ \midrule
Original Token  & 36.66     & 29.88    & 28.26 & 66.95  & 53.74 & 42.34 & -        & -        & -      \\ \midrule
LLMLingua-2-small*        & 31.55 & 29.04 & 26.79 & 49.24 & 32.61 & 33.94        & 1.61$\times$ & 1.16$\times$        & 1.12$\times$                    \\
LLMLingua-2* & 31.35     & \textbf{30.57}    & \textbf{27.29} & 52.29  & 33.87 & 35.16   & 1.61$\times$ & 0.83$\times$   & 0.97$\times$ \\
SDTP   & \textbf{32.70}      & 27.74    & 26.65 & \textbf{62.74}  & \textbf{53.99} & \textbf{39.82}   & 1.61$\times$  & \textbf{1.71$\times$}   & \textbf{1.31$\times$} \\ \midrule
SDTP-low   & 35.60         & 28.83      & 27.26        & 64.58      & 52.43      & 40.98         & 1.32$\times$        & 1.37$\times$        & 1.10$\times$     \\ 
\bottomrule
\end{tabular}}
\end{table}

\paragraph{Inference latency and FLOPs.}
In order to analyze the performance gains under different sequence lengths in detail, a series of speed tests were conducted using specifically constructed data. In these tests, the prompt of the constructed data is padded to the same length, and the LLMs are required to generate tokens of the same length, which is set to 128 here. The context lengths of the prompts ranged from 4K to 128K. The theoretical maximum reduction in end-to-end FLOPs after the integration of the SDTP module was predicted to be 47.2\%. The end-to-end and prefill phase inference latency were recorded for various prompt lengths. Additionally, the peak GPU memory usage during inference was monitored. 

The results presented in Table~\ref{tab:flops-speedup-gpu} indicate that the SDTP module significantly reduces inference latency, achieving an end-to-end speedup of up to 1.75 times. This speedup is substantial, demonstrating the effectiveness of the SDTP module in optimizing model performance. Furthermore, the SDTP module also effectively reduced the GPU memory required during inference, leading to a memory saving of up to 34.26\%. This reduction in memory usage is particularly beneficial for large-scale applications with limited computational resources. 

Overall, the findings from these experiments confirm that the SDTP module not only enhances the speed of inference but also reduces the memory footprint of LLMs, making them more suitable for deployment on systems with constrained resources.

\begin{table}[t]
\caption{Inference latency, FLOPs and GPU memory usage test on Mistral-7B with context lengths from 4K to 128. Our SDTP can achieve up to 47.2\% FLOPs reduction, 1.75$\times$ end-to-end acceleration, and 34.26\% memory savings.}
\label{tab:flops-speedup-gpu}
\small
\centering
\begin{tabular}{lc|cc|cc|c}
\toprule
\multirow{2}{*}{\begin{tabular}[c]{@{}l@{}}Seq.\\ length\end{tabular}} & \multirow{2}{*}{Method} & \multicolumn{2}{c|}{TFLOPs} & \multicolumn{2}{c|}{Latency(s)} & \multirow{2}{*}{Memory(GB)} \\ 
\cmidrule{3-6}
                      &            & Prefill  & E2E      & Prefill & End2End    &        \\ 
\midrule
\multirow{2}{*}{4K}   & $w/o$ SDTP & 72.51   & 74.78   & 0.31    & 3.70    & 15.55 \\ \cmidrule{2-7} 
                      & $w$ SDTP   & 47.81   & 50.01   & 0.20($\uparrow 1.52\times$)    & 3.43($\uparrow 1.08\times$)    & 15.04($\downarrow 3.28\%$) \\ \midrule
\multirow{2}{*}{8K}   & $w/o$ SDTP & 158.22  & 160.70  & 0.64    & 4.85    & 17.10 \\ \cmidrule{2-7} 
                      & $w$ SDTP   & 102.01  & 104.35  & 0.40($\uparrow 1.58\times$)    & 3.98($\uparrow 1.22\times$)    & 15.86($\downarrow 7.25\%$)  \\ \midrule
\multirow{2}{*}{16K}  & $w/o$ SDTP & 369.22  & 372.11  & 1.40    & 7.17    & 20.20 \\ \cmidrule{2-7} 
                      & $w$ SDTP   & 229.41  & 232.03  & 0.84($\uparrow 1.67\times$)    & 5.46($\uparrow 1.31\times$)    & 17.55($\downarrow 13.11\%$) \\ \midrule
\multirow{2}{*}{32K}  & $w/o$ SDTP & 949.54  & 953.25  & 3.40    & 13.63   & 26.41 \\ \cmidrule{2-7} 
                      & $w$ SDTP   & 560.40  & 563.56  & 1.90($\uparrow 1.79\times$)    & 8.95($\uparrow 1.52\times$)    & 20.95($\downarrow 20.67\%$) \\ \midrule
\multirow{2}{*}{64K}  & $w/o$ SDTP & 2743.46 & 2748.83 & 9.33    & 27.75   & 39.31 \\ \cmidrule{2-7} 
                      & $w$ SDTP   & 1526.98 & 1531.25 & 4.99($\uparrow 1.87\times$)    & 16.92($\uparrow 1.64\times$)   & 28.24($\downarrow 28.18\%$) \\ \midrule
\multirow{2}{*}{128K} & $w/o$ SDTP & 8864.49 & 8873.15 & 29.30   & 63.69   & 65.13  \\ \cmidrule{2-7} 
                      & $w$ SDTP   & 4678.81 & 4685.28 & 13.60($\uparrow 2.15\times$)   & 36.35($\uparrow 1.75\times$)   & 42.82($\downarrow 34.26\%$) \\ 
\bottomrule
\end{tabular}
\end{table}

\subsection{Generalization of SDTP}
\paragraph{Generalization of SDTP on different tasks.}Since our method freezes all Transformer blocks and only inserts dynamic token pruning modules between blocks, our method can theoretically preserve the generalization of the original LLM model. We have verified this through experiments on three LLMs with eight 5-shot downstream tasks: COPA~\cite{roemmele2011choice}, PIQA~\cite{bisk2020piqa}, Winogrande~\cite{sakaguchi2021winogrande}, MathQA~\cite{amini2019mathqa}, BoolQ~\cite{clark2019boolq}, CB~\cite{roberts2020much}, WiC~\cite{pilehvar2018wic}, and WSC~\cite{kocijan2020review}. As shown in Table~\ref{llm-sdtp-generalization-table}, the model retains comparable performance to the original model after cumulatively pruning up to 65\% of tokens.
\begin{table}[htbp]
\caption{Accuracy of eight 5-shot downstream tasks on three LLMs: Llama2-7B, Mistral-7B, and BLOOM-7B. Our SDTP is generalizable to various models and tasks by hierarchically pruning 65\% of the input tokens.}
\label{llm-sdtp-generalization-table}
\centering
\begin{tabular}{l|cc|cc|cc}
\toprule
\multirow{2}{*}{Method} & \multicolumn{2}{c|}{Llama2-7B} & \multicolumn{2}{c|}{Mitral-7B} & \multicolumn{2}{c}{BLOOM-7B} \\
\cmidrule(l){2-7}
              & Full    & SDTP    & Full    & SDTP    & Full    & SDTP    \\ \midrule
COPA          & 83.00 & 85.00 & 90.00 & 91.00 & 69.00 & 69.00 \\
PIQA          & 78.84 & 78.89 & 84.71 & 84.44 & 73.61 & 71.60 \\
WinoGrande    & 73.88 & 72.22 & 76.64 & 76.16 & 65.59 & 60.85 \\
MathQA        & 31.93 & 31.19 & 36.45 & 36.78 & 27.81 & 27.37 \\
BoolQ         & 60.98 & 66.42 & 66.91 & 71.35 & 61.19 & 61.77 \\
CB            & 51.79 & 53.57 & 92.86 & 89.29 & 48.21 & 53.57 \\
WiC           & 52.04 & 55.17 & 57.21 & 55.49 & 50.31 & 50.47 \\
WSC           & 36.54 & 36.54 & 66.35 & 65.38 & 36.54 & 36.54 \\ \midrule
Avg           & 58.62 & 59.88 & 71.39 & 71.24 & 54.03 & 53.90 \\ \midrule
Pruning ratio & -        & 65\%    & -        & 65\%    & -        & 65\%   \\
\bottomrule
\end{tabular}
\end{table}
\paragraph{Results of different LLMs.}
As shown in Table~\ref{llm-sdtp-generalization-table}, we performed experiments on three Transformer-based LLMs with 7 billion parameters: Llama2-7B, Mistral-7B, and BLOOM-7B. The results demonstrate that our approach achieves accuracy comparable to all three original, unpruned LLMs. This finding suggests the versatility of the SDTP framework, indicating its potential for integration into various Transformer-based LLMs without sacrificing performance.
\subsection{Ablation Results}
\paragraph{Ablation on the saliency-driven dynamic token pruning.}
For comparison, we implemented a token pruning module with only token pruning rate constraints similar to DynamicVit~\cite{rao2021dynamicvit}, and trained it on the Dolly dataset to make it suitable for NLP tasks as our baseline. We test the effectiveness of our proposed saliency-driven token selection module and ranking-based optimization strategy using two target LLM models: Llama2-7B and Mistral-7B. As illustrated in Table~\ref{ablation-llama2-table}, we compared four 5-shot evaluation tasks: COPA, PIQA, Winogrande, MathQA. The results demonstrate that our proposed saliency-driven dynamic token pruning module effectively retains crucial tokens and that pairwise ranking loss further aids the training of the token pruning module. Our SDTP can improve the accuracy of downstream tasks with an improvement of 0.89 on average compared to the baseline.
As shown in Table~\ref{ablation-mistral-table}, we compared five types of tasks, such as single-document QA and multi-document QA on the multi-task long context understanding benchmark LongBench~\cite{bai2023longbench}. The results demonstrate that our SDTP can improve the accuracy of long context understanding tasks, especially for multi-document QA tasks, which can be improved by 1.12. 
\begin{table}[]
\caption{Ablation on four 5-shot tasks with Llama2-7B as target LLM. Our proposed saliency-driven token selection module and ranking-based optimization strategy can improve the accuracy of downstream tasks, with an improvement of 0.89 on average compared to the baseline.}
\label{ablation-llama2-table}
\centering
\resizebox{\linewidth}{!}{
\begin{tabular}{lccccc}
\toprule
Method    & COPA    & PIQA    & Winogrande & MathQA & Average  \\ 
\midrule
SDTP & \textbf{85.00} & \textbf{78.89} & \textbf{72.22}    & \textbf{31.19}    & \textbf{66.82} \\
$w/o$ Rank     & 84.00(-1.00) & 78.40(-0.49) & 71.82(-0.40)    & 30.85(-0.34)    & 66.27(-0.55) \\
$w/o$ Rank \& MSE  & 83.00(-2.00) & 78.29(-0.60) & 71.43(-0.79)    & 30.99(-0.20)    & 65.93(-0.89) \\
\bottomrule
\end{tabular}}
\end{table}

\begin{table}[htbp]
\caption{Ablation on LongBench with Mistral-7B as target LLM. Our proposed saliency-driven token selection module and ranking-based optimization strategy can improve the accuracy of long context understanding tasks. Especially on multi-document QA tasks, it can be improved by 1.12.}
\label{ablation-mistral-table}
\centering
\resizebox{\linewidth}{!}{
\begin{tabular}{lccccc}
\toprule
Method & SingleDoc & MultiDoc & Summ. & FewShot & Code \\
\midrule
SDTP & \textbf{32.65} & \textbf{27.28} & \textbf{26.60} & \textbf{62.63} & \textbf{53.81} \\
$w/o$ Rank     & 32.23(-0.22) & 26.51(-0.77) & 26.32(-0.28) & 61.91(-0.72) & 53.34(-0.47)  \\
$w/o$ Rank \& MSE  & 31.94(-0.71) & 26.16(-1.12) & 26.59(-0.01) & 62.00(-0.63) & 53.59(-0.22)  \\
\bottomrule
\end{tabular}}
\end{table}

\subsection{Compatible with KV Cache Compression}
The token pruning method is designed to reduce the number of FLOPs during the prefill phase, while the KV cache compression method aims to decrease FLOPs in the decode phase. Since these two methods target different stages of the computation and do not interfere with each other, they are considered orthogonal. In the experiment described, the compatibility of the proposed method with KV cache compression was tested with Llama2-7B. The $\rm H_2O$~\cite{zhang2024h2o} algorithm was employed for this purpose. During the KV cache storage phase, $\rm H_2O$ compresses the KV cache, reserving 40\% of it, which can lead to further improvements in inference speed and throughput.
Table~\ref{compatibility-h2o-table} indicates that the proposed method is compatible with KV cache compression, as it does not introduce compound errors. This means that the benefits of both techniques reduced FLOPs from token pruning and efficient cache management from compression can be simultaneously realized without negatively affecting the accuracy or performance of the model.

\begin{table}[]
\caption{Compatibility with KV cache compression with Llama2-7B as target LLM. Our SDTP is compatible with KV cache compression and does not introduce compound errors by hierarchically pruning 65\% of the input tokens.}
\label{compatibility-h2o-table}
\centering
\begin{tabular}{l|c|ccc}
\toprule
\multirow{2}{*}{Tasks} & \multirow{2}{*}{Full} & \multicolumn{3}{c}{40\% KV}  \\
                       &                       & Local   & $\rm H_2O$     & SDTP $w. \rm H_2O$ \\ \midrule
COPA                   & 83.00               & 50.00 & 85.00 & 85.00  \\
PIQA                   & 78.84               & 50.27 & 78.62 & 78.56  \\
Winogrande             & 73.88               & 48.46 & 73.32 & 72.53  \\
MathQA                 & 31.93               & 23.38 & 30.99 & 30.99  \\ \midrule
Pruning ratio          & -                      & -        & -        & 65\%    \\
\bottomrule
\end{tabular}
\end{table}

\subsection{Experiment on Pruning Rate and Pruning Strategy}
\begin{figure}[htbp]
\centering
  \begin{minipage}[b]{0.48\textwidth}
  \centering
    \includegraphics[height=4cm]{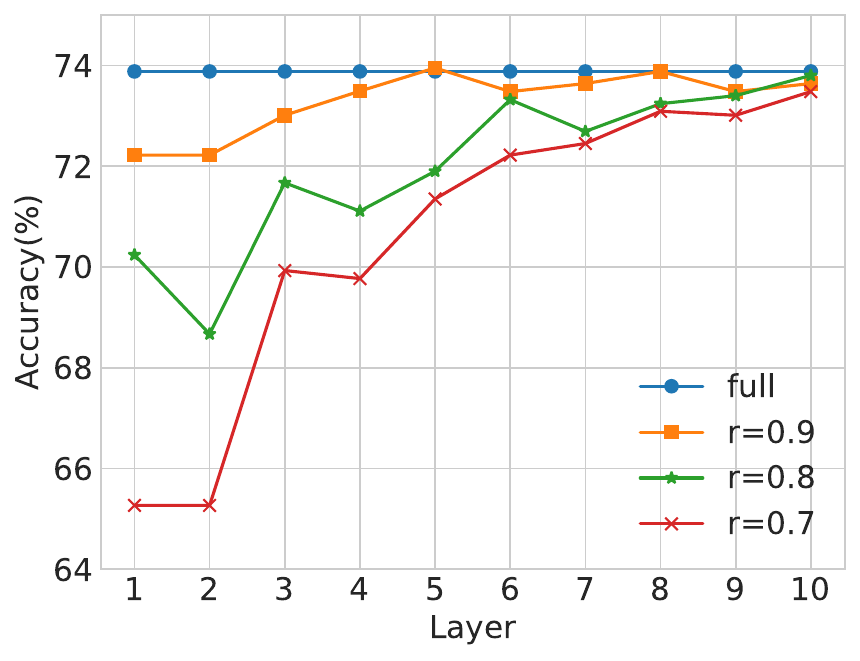}
    \caption{Effect of first pruning layer.}
    \label{fig:pruning_rate_layer_add}
  \end{minipage}
  \hfill
  \begin{minipage}[b]{0.48\textwidth}
  \centering
    \includegraphics[height=4cm]{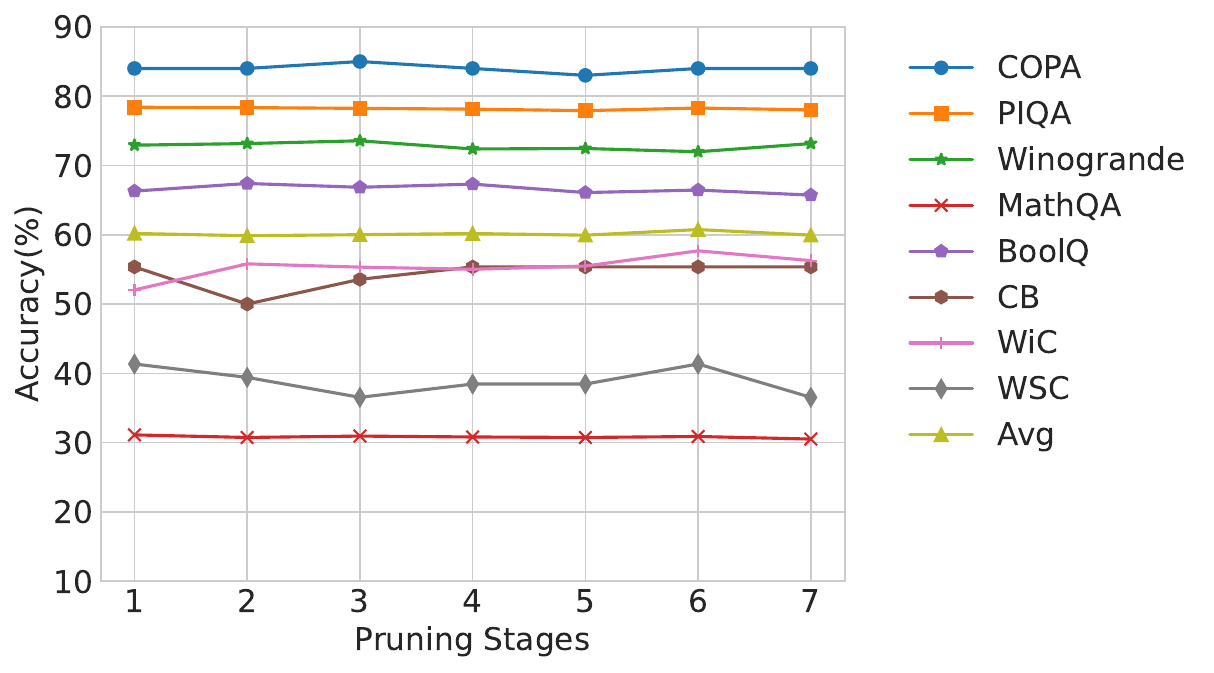}
    \caption{Effect of multi-stage pruning.}
    \label{fig:pruning_stages}
  \end{minipage}
\end{figure}

To optimize the token pruning strategy for minimal impact on model performance, a series of experiments were conducted using Llama2-7B as the target LLM on the 5-shot task Winogrande. These experiments involved single-stage pruning from the first layer up to the tenth layer, with varying keeping ratios of 90\%, 80\%, and 70\%. The goal was to identify the optimal layer to start pruning and the ideal keeping ratio that would least affect the model's performance. As shown in Figure~\ref{fig:pruning_rate_layer_add}, it was observed that starting the pruning process from the fourth layer with a conservative keeping ratio of 90\% can achieve a balance between model efficiency and performance maintenance. This finding suggests that earlier layers are less amenable to aggressive pruning. Conversely, later layers, which tend to be more specialized and may contain more redundant or less critical information, can tolerate higher levels of pruning without significant degradation in performance. This outcome is important for designing efficient pruning strategies that reduce computational overhead without compromising accuracy. 

To further investigate the effect of multi-stage pruning on model accuracy, experiments were conducted across eight 5-shot tasks. In these experiments, a keeping ratio of 90\% was maintained while pruning began from the fourth layer, with a layer step of 3. The layer step refers to the number of layers between each successive pruning stage. As shown in Figure~\ref{fig:pruning_stages}, the results indicated that although a high pruning rate in a single stage could be detrimental to model performance, increasing the number of pruning stages had a minor impact on performance. This is in line with the earlier observation in the article regarding the increase in token sparseness with deeper layers. The rule suggests that significant token sparseness occurs as you move deeper into the layers of LLM, allowing for more aggressive pruning in later stages without substantially affecting the model's ability to process and generate language.

\section{Conclusion}
\label{sec:con}
This work introduces a novel approach to accelerate LLM inference through dynamic redundant token pruning. We propose the Saliency-Driven Token Pruning (SDTP) module, a lightweight and adaptable component that seamlessly integrates within the Transformer layer. The SDTP module leverages a supervised training strategy guided by token saliency and a ranking-based optimization strategy to effectively select tokens for pruning during inference. Our experimental results demonstrate that SDTP achieves significant inference acceleration, progressively pruning up to 65\% of input tokens while attaining a speedup of 1.75$\times$. Notably, this method is applicable to a wide range of mainstream Transformer-based LLMs and offers the potential for further compression when combined with KV cache compression techniques.

\bibliographystyle{plainnat}
\bibliography{neurips_2024}

\end{document}